\documentclass{article}

\usepackage[final]{neurips_2024}

\def\EAAI{0}

\usepackage[utf8]{inputenc} 
\usepackage[T1]{fontenc}    
\usepackage{url}            
\usepackage{booktabs}       
\usepackage{amsfonts}       
\usepackage{nicefrac}       
\usepackage{microtype}      
\usepackage{xcolor}         

\usepackage{todonotes}

\usepackage{algorithm}
\usepackage{algorithmic}

\usepackage{newfloat}
\usepackage{listings}
\floatstyle{ruled}
\newfloat{listing}{tb}{lst}{}
\floatname{listing}{Listing}
\usepackage{amssymb}
\usepackage{amsmath}
\usepackage{booktabs}
\usepackage{dirtytalk}
\usepackage{dsfont}
\usepackage{mathtools}
\usepackage{mathrsfs}
\usepackage{subcaption}
\usepackage{tcolorbox}

\def\A{\mathcal{A}}
\def\E{\mathcal{E}}
\def\Epref{\E_{\textrm{pref}}}
\def\Eprereq{\E_{\textrm{prereq}}}
\def\V{\mathcal{V}}
\def\C{\mathcal{C}}

\def\x{\mathbf{x}}
\def\s{\mathbf{s}}

\def\ot{\mathbf{o}_t}

\def\W{\mathbf{W}}

\usepackage[colorlinks,citecolor=blue]{hyperref}       

\title{A Pre-Trained Graph-Based Model for Adaptive Sequencing of Educational Documents}

\author{
    Jean Vassoyan\textsuperscript{1,2} \quad
    Anan Schütt\textsuperscript{3} \quad
    Jill-Jênn Vie\textsuperscript{4} \quad
    Arun-Balajiee Lekshmi-Narayanan\textsuperscript{5}\\
    \textbf{Elisabeth André\textsuperscript{3}} \quad
    \textbf{Nicolas Vayatis\textsuperscript{1}}\\
    \vspace{0.5em} 
    \\
    \textnormal{\textsuperscript{1}Université Paris-Saclay, CNRS, ENS Paris-Saclay, Centre Borelli, France} \\
    \textnormal{\textsuperscript{2}onepoint, France} \quad
    \textnormal{\textsuperscript{3}University of Augsburg, Germany} \quad
    \textnormal{\textsuperscript{4}Soda Team, Inria Saclay, France}\\
    \textnormal{\textsuperscript{5}University of Pittsburgh, USA}
    \\
    \vspace{0.5em} 
    \\
    \textnormal{\texttt{\{jean.vassoyan, nicolas.vayatis\}@ens-paris-saclay.fr}}\\
    \texttt{\{anan.schuett, elisabeth.andre\}@uni-a.de} \\ \texttt{arl122@pitt.edu} \quad \texttt{jill-jenn.vie@inria.fr} 
}

\begin{document}

\maketitle

\begin{abstract}
Massive Open Online Courses (MOOCs) have greatly contributed to making education more accessible.
However, many MOOCs maintain a rigid, one-size-fits-all structure that fails to address the diverse needs and backgrounds of individual learners.
Learning path personalization aims to address this limitation, by tailoring sequences of educational content to optimize individual student learning outcomes.
Existing approaches, however, often require either massive student interaction data or extensive expert annotation, limiting their broad application.
In this study, we introduce a novel data-efficient framework for learning path personalization that operates without expert annotation.
Our method employs a flexible recommender system pre-trained with reinforcement learning on a dataset of raw course materials.
Through experiments on semi-synthetic data, we show that this pre-training stage substantially improves data-efficiency in a range of adaptive learning scenarios featuring new educational materials.
This opens up new perspectives for the design of foundation models for adaptive learning.
\end{abstract}



\section{Introduction}


One-on-one tutoring has been shown to yield higher learning gains than one-to-many teaching \citep{bloom19842}, encouraging the development of \emph{adaptive learning} as a research area, with a wide range of open problems.
Learning path personalization is one of them \citep{brusilovsky2007user}: given an e-learning platform with a corpus of educational documents, how can we recommend a sequence of these documents to any student in order to maximize individual learning gains? Personalization is common in computerized adaptive testing, where the questions are chosen on-the-fly based on the performance of the examinee. Knowledge tracing, the task of modeling the acquisition of knowledge, has been done either explicitly using graphical models~\citep{corbett1994knowledge,doignon1985spaces,leighton2004attribute}, or implicitly using neural networks~\citep{NIPS2015bac9162b,ghosh2020context}.


The problem of learning path personalization can be formulated as a Markov decision process (MDP), where each episode is a learning session with a student, each action is a recommendation (of document) and the reward signal is the learning gains of the student.
Consequently, many attempts have been made to tackle this problem with reinforcement learning (RL) algorithms \citep{clement2015multi, bassen2020reinforcement, shabana2022curriculumtutor}.
However, collecting human learning data is expensive, making these approaches very sensitive to the well-known problem of sample efficiency.
Therefore, standard deep RL approaches are usually impractical, as one can hardly afford to make a model interact with thousands of humans.

\ifx\EAAI\undefined

As
\else

\begin{figure}[!t]
    \centering
    \includegraphics[width=0.6\columnwidth]{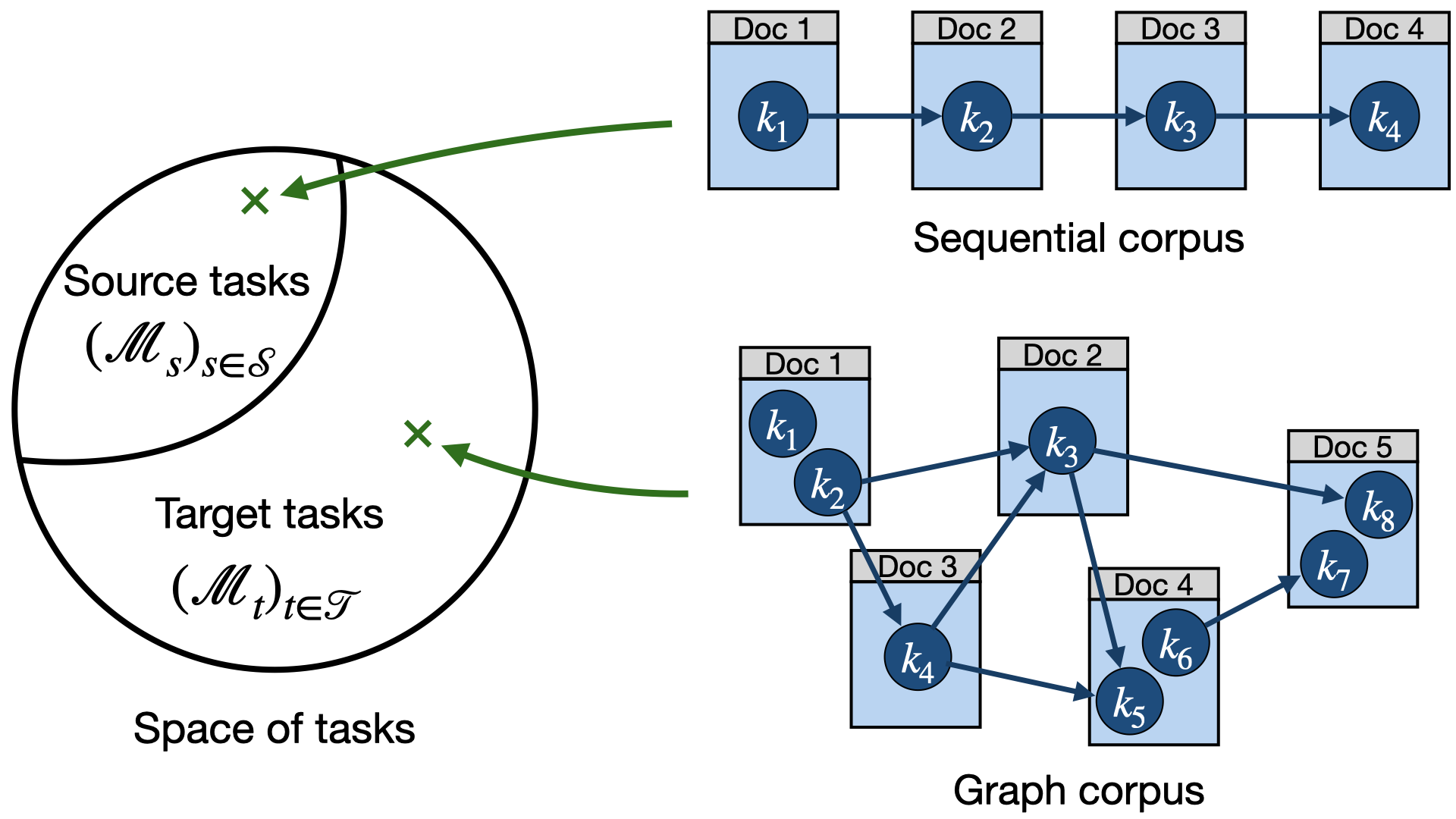}
    \caption{Illustration of the partition of the domain: \textit{sequential} corpora are designed to be followed in one single way whereas \textit{graph} corpora can be navigated along a variety of paths. This dichotomy can be used to pre-train a recommender system on readily available source tasks $(\mathcal{M}_s)_{s \in \mathcal{S}}$ and fine-tune it on a target task ${\mathcal{M}}_t$.}
    \label{fig:sequential_graph_corpora}
\end{figure}

This falls under a common issue in applied RL where the environment is hardly accessible for training.
One prevalent approach to addressing this issue is to train the RL agent in a simulated environment close enough to \say{real-world} conditions \citep{towers_gymnasium_2023, tassa2018deepmind}.
However, simulating student behavior generally requires extensive annotation of educational activities with knowledge components and prerequisite relationships \citep{corbett1994knowledge,thaker2020recommending}.
These are costly to extract and rarely available in adaptive learning platforms.
Another common approach is offline RL \citep{lange2012batch, levine2020offline}, which allows training an RL agent on real-world educational data, alleviating the dependence on a specific student simulator.
However, the lack of widely available data in adaptive learning environments complicates this approach.
Another solution can be found in transfer learning: if
\fi
the target environment is not easily accessible, one can leverage information from similar - yet more accessible - environments to pre-train the agent.
The rich literature on transfer learning in deep RL has been covered by \citet{zhu2023transfer} and numerous works have demonstrated the benefit of transferring representations from source to target domains \citep{rusu2016progressive, devin2017learning, zhang2018}.
In this paper, we propose to apply such transfer learning strategy to the problem of learning path personalization.
Our approach relies on a partition of the domain between two sets of tasks: the \textit{source} and \textit{target} tasks (see Figure \ref{fig:sequential_graph_corpora}).
The source tasks, referred to as \textit{sequential corpora}, are corpora designed for sequential navigation, as is commonly the case in MOOCs.
These are non-adaptive but represent massive amounts of data.
The target tasks, referred to as \textit{graph corpora}, denote corpora that can be browsed in multiple ways: these are more suitable for adaptive learning but represent a smaller amount of data.
Our contribution is twofold. We first set up a mathematical framework for modeling the behavior of a student population interacting with a distribution of educational corpora.
This framework enables to translate the dichotomy between \emph{sequence} and \emph{graph} corpora into a partition of the domain between \emph{source} and \emph{target} tasks.
Then we show experimentally that a recommender system pretrained on a set of educational corpora (source tasks) can transfer its knowledge to a brand new adaptive learning corpus (target task) and benefit from a considerable increase in sample efficiency, especially in the small data regime.
To the best of our knowledge, this is the first pre-trained recommender system for learning path personalization. \footnote{Our code and experiments are available at:\\ \url{https://github.com/jvasso/pretrain-rl-adaptive-learning}.}

\section{Formalization of the problem}
\label{sec:formulation}

Given a set of educational corpora $\mathscr{C}$ and a student population $\mathcal{P}$, we formalize the problem of learning path personalization as a set of POMDPs.
Each episode starts by sampling a corpus $\mathcal{C}$ from $\mathscr{C}$ and a student $u$ from $\mathcal{P}$.
The task consists in recommending a sequence of documents from $\mathcal{C}$ to maximize the learning gains of student $u$.
We first model the joint distribution over $(u,\mathcal{C})$ as a random graph, then we show how it can be used to formalize the problem as a collection of POMDPs that can be partitioned into \textit{source} and \textit{target} tasks.

\subsection{Student and corpus distributions}


In the literature on adaptive learning, a corpus $\mathcal{C}$ of educational materials is usually characterized by three elements \citep{thaker2020recommending, leighton2004attribute}:
\begin{itemize}
    \item a set of documents $\mathcal{D}=\{d_1, d_2, \dots\}$;
    \item a set of knowledge components $\mathcal{V} = \{k_1, k_2, \dots\}$; a knowledge component (KC) is a small unit of knowledge taught by a document ; we denote $d_{\rightarrow} \subset \mathcal{V}$ the set of KCs taught by document $d$;
    \item a set of prerequisite relationships $\Eprereq \subset \mathcal{V}^2$ between these KCs ; $(k_i \to k_j)\in \Eprereq$ means that $k_i$ is a prerequisite for $k_j$.
\end{itemize}



On the other hand, the behavior of a student $u$ interacting with $\mathcal{C}$ is determined by these main components: 
\begin{itemize}
    \item his prior knowledge $\mathbf{x}^{(u)} \in \{0,1\}^{|\mathcal{V}|}$: this is a binary vector stating which KCs were already known by the student prior to the learning session;
    \item his learning preferences $\mathcal{E}_{\text{pref}}^{(u)} \subset \mathcal{V}^2$: these are similar to the prerequisite relationships $\mathcal{E}_{\text{prereq}}$ but depend on the student\footnote{this choice of modeling learning preferences as an additional set of prerequisite relationships is a particularity of our modeling. It is further motivated in Appendix \ref{appendix:learning_prefs}.};
    we denote \mbox{$\mathcal{E}^{(u)} = \mathcal{E}_{\text{prereq}} \cup \mathcal{E}_{\text{pref}}^{(u)}$}; we also denote $d_\leftarrow^{(u)} \subset \mathcal{V}$ the requirements of document $d$ for student $u$: $d_\leftarrow^{(u)} = \underset{k\in d_\rightarrow}\bigcup \{ k^\prime | (k^\prime \rightarrow k) \in \mathcal{E}^{(u)} \}$;
    \item his ability to demonstrate his knowledge when the opportunity arises; this is typically modeled as a function that maps a knowledge state $s$ and a document $d$ to an observation $o$ of the student's knowledge 
    , through a likelihood function: $\mathcal{Z}(s, d, o)=\mathbb{P}(\mathbf{o}_t=o\mid \mathbf{s}_{t}=s, \mathbf{d}_t=d)$;
    we assume that $\mathcal{Z}$ is identical for all students.
\end{itemize}

\begin{figure*}[t]
    \centering
    \includegraphics[width=0.99\textwidth]{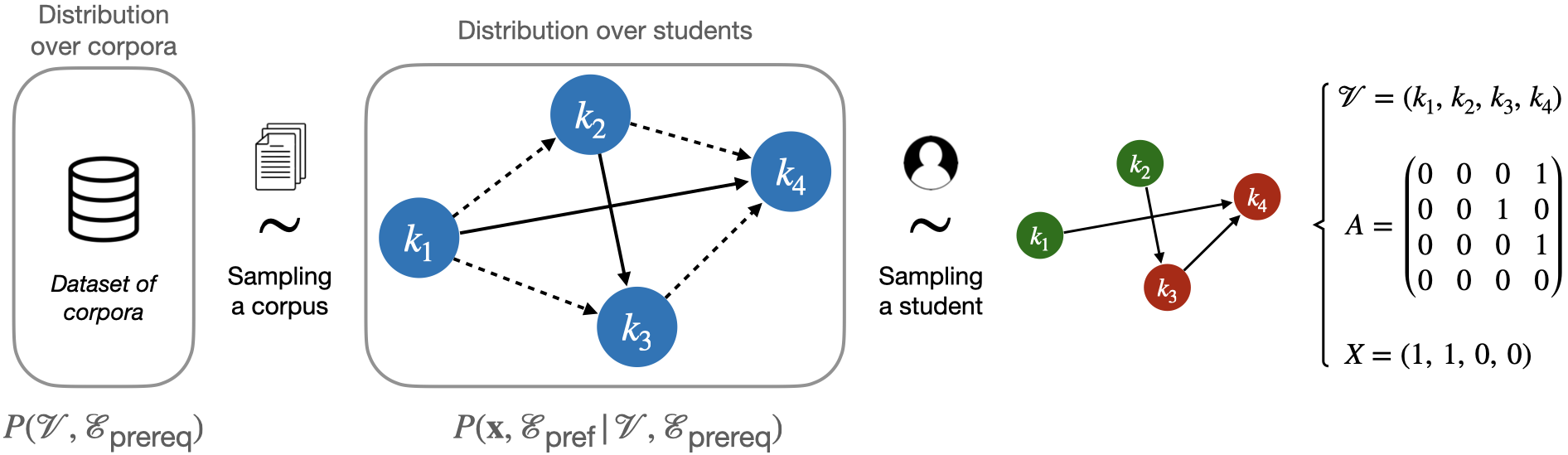}
    \caption{Illustration of the sampling process of a student-corpus pair. $A$ is the adjacency matrix corresponding to the set of edges $\mathcal{E}=\mathcal{E}_{\text{prereq}}\cup \mathcal{E}_{\text{pref}}$.}
    \label{fig:student_population}
\end{figure*}

We drop superscripts $(u)$ in the following.
All previously defined variables play a role in the sampling of a student-corpus pair $(u, \mathcal{C})$.
However, in our probabilistic modeling, we only write the key variables $(\mathcal{V}, \mathcal{E}, \mathbf{x})$, as they are the only sources of variability among students.
Consequently, a student-corpus pair $(u, \mathcal{C})$ can be modeled as a graph \mbox{$G = \bigl(\mathcal{V}, \mathcal{E}, \mathbf{x} \bigr)$}
and the joint distribution over $(u, \mathcal{C})$ can be expressed as a random graph \mbox{$\mathcal{G} \sim P\bigl(\mathcal{V}, \mathcal{E}, \mathbf{x} \bigr)$}.
Note that $\mathcal{E}$ and $\mathbf{x}$ are not independent since any knowledge state has to be consistent with the requirements. 
In practice, we carry out the sampling process in two steps: sampling a corpus, then sampling a student (see Figure \ref{fig:student_population}).
\begin{equation}
    P(u, \mathcal{C}) = P(u|\mathcal{C}) P(\mathcal{C})
    = P(\mathcal{E}_{\text{pref}}, \mathbf{x} | \mathcal{V}, \mathcal{E}_{\text{prereq}}) P(\mathcal{V}, \mathcal{E}_{\text{prereq}}) \notag
\end{equation}


\subsection{Definition of the domain}

We now define the domain, i.e. the set of tasks, as a collection of POMDPs. Given a corpus $\C = (\V,\Eprereq)$, a recommendation task is a POMDP \mbox{$\mathcal{M}(\C) = (\mathcal{S},\A,\mathcal{O},\mathcal{T},\mathcal{R},\mathcal{Z})$}.
$\mathcal{S} \subset \{0,1\}^{|\mathcal{V}|} \times \mathcal{V}^2$ is the state space, which means that each student state $\mathbf{s}_t \coloneqq (\mathbf{x}_t, \Epref)$ is defined as a combination of the student's knowledge and learning preferences (we assume $\Epref$ remains constant throughout the episode).
$\mathbf{a}_t\in \mathcal{A}$ is the document recommended at step $t$ (also denoted $\mathbf{d}_t$).
$\mathbf{o}_t\in \mathcal{O}$ is an observation of the student's knowledge at time $t$: in our setting, $\ot$ is a feedback signal qualifying the student's understanding of $\mathbf{d}_t$.
This feedback can take 3 possible values: $\ot=f_<$ if the student did not understand the document (because he didn't have the prerequisites) ; $\ot=f_\circ$ if he understood the document and learned something new ; $\ot=f_>$ if he understood the document but did not learn anything new (he already knew all the KCs taught by the document).



This is formalized by the observation function:
\begin{equation}
    \mathcal{Z}(s, d, o) ={P}(\mathbf{o}_{t}={o}\mid \s_{t}={s}, \mathbf{a}_t={d}) =
\begin{cases}
    1 - m_s(d_\leftarrow), & \text{if } o = f_< \ \textrm{ (too hard)}\\
    m_s(d_\rightarrow), & \text{if}\  o = f_> \ \textrm{ (too easy})\\
    m_s(d_\leftarrow) \bigl(1 - m_s(d_\rightarrow) \bigr), & \text{if}\  o = f_\circ  \ \textrm{ (right level})\\
\end{cases}
\end{equation}
where $s = (\x, \Epref)$ is the student state and $m_s(\cdot)$ is a function that states whether a student in state $s$ masters a given set of KCs: $\ m_s(\mathbf{v}) = \bigcap_{k_j \in \mathbf{v}}\{ x_j=1 \}$.
Note that $m_s(d_\leftarrow)$ and $1 - m_s(d_\rightarrow)$ cannot simultaneously be zero, as $d_\leftarrow$ is always a prerequisite of $d_\rightarrow$, making the likelihood function valid.



The transition probability function $\mathcal{T}(s,d,{s}^{\prime})$, unknown to the recommender system, simply states that a student returning feedback $f_\circ$ on document $d$ will learn all the knowledge components taught by $d$.
Otherwise the student will remain in the same knowledge state.
More formally:
\begin{equation}
   \mathcal{T}(s,d,{s}^{\prime}) = \mathbb{P}(\mathbf{s}_{t+1}={s}^{\prime}\mid \mathbf{s}_t={s},\mathbf{a}_t={d}) = \prod_{k_i \in \mathcal{V}} \tau(s, d, x_i^\prime)
\end{equation}
where $s = (\x, \Epref)$, $s' = (\x', \Epref)$ and $\forall i \in \{1, \ldots, |\mathcal{V}|\}$:
\begin{align}
\label{eq:transition-short}
    \tau(s, d, x_i^\prime) =
\begin{cases}
    \mathds{1}_{\{k_i\in d_\rightarrow\}} (1 - m_s(d_\leftarrow)) + \mathds{1}_{\{k_i \notin d_\rightarrow \}} &
    \text{if} \ x_i=x_i^\prime \\
    \mathds{1}_{\{(k_i\in d_\rightarrow) \wedge (x_i^\prime=1) \}} m_s(d_\leftarrow) &
    \text{if} \ x_i=1-x_i^\prime
\end{cases}
\end{align}





A derivation of this formula is provided in Appendix \ref{appendix:transition-probability}. The reward function $\mathcal{R}(\mathbf{s}_t,\mathbf{s}_{t+1})$, is the learning gains of the student from steps $t$ to step $t+1$, i.e. the number of newly acquired KCs:
    \begin{eqnarray}\label{eq:reward}
        \mathcal{R}(\mathbf{s}_t,\mathbf{s}_{t+1}) = |\mathbf{x}_{t+1} - \mathbf{x}_{t}|
    \end{eqnarray}
\noindent
where  $\s_{t + 1} = (\x_{t + 1}, \Epref)$, $\s_{t} = (\x_t, \Epref)$, and $|\cdot|$ is the $L^1$ norm.
As in most reinforcement learning problems, the goal of the above POMDP $\mathcal{M}(\C)$ is to find a policy \mbox{$\pi :\mathcal{O}\rightarrow \mathcal{A}$} that maximizes the expected return over each episode: \mbox{$\pi^* = \underset{\pi}{\arg \max }\ \mathbb{E} \left[\  \sum_{t=0}^{T-1} \gamma ^t \mathcal{R}(\mathbf{s}_t,\mathbf{s}_{t+1}) \ \right]$}.

\subsection{Partition of the domain}\label{subsec:seq_vs_graph_corpora}

We split the space of corpora into two subspaces: \textit{sequential} corpora and \textit{graph} corpora (see Figure \ref{fig:sequential_graph_corpora}).
On the one hand, \textit{sequential} corpora relate to \say{standard} courses, designed to be fully completed, following an identical path for each student.
We argue that the set of sequential corpora is a great source of pre-training data for three reasons.
First, they represent a huge amount of data (most e-learning curricula are designed that way).
Second, since they are designed to be followed sequentially, we consider that each corpus can be reasonably approximated with a chain of KCs where each document teaches one KC (this is reasonable if the documents are small enough): for $\mathcal{C} = (\mathcal{V}, \mathcal{E}_{\text{prereq}})$ with $\V = \{k_1, \ldots, k_n\}$, we have $\mathcal{E}_{\text{prereq}}= \{(k_i \to k_{i + 1})\}_{1 \leq i < n}$ and $d_{i \rightarrow} = \{ k_i \}$ for all $i = 1, \ldots, n$.
This allows to bypass the extensive tagging process usually done by experts.
Third, even though it does not leave much room for adaptivity (we have necessarily $\mathcal{E}_{\text{pref}}=\varnothing$), the structure of a sequential course, designed by an expert of the domain, is a great source of information on the relationship among technical concepts.
Consequently, we refer to the set of sequential corpora as the set of source tasks, denoted $\boldsymbol{\mathcal{M}}_s$.

On the other hand, \textit{graph} corpora are much more flexible: some knowledge components can be reached through multiple paths, depending on the student's background and learning preferences.
In this case, $\C$ is a non-sequential graph, and $\Epref$ is not empty.
These corpora are much more suited for adaptive learning and, therefore, correspond to the real purpose of our recommendation engine.
We call them target tasks and denote $\boldsymbol{\mathcal{M}}_t$ the corresponding set.

Following the formalism of \citet{zhu2023transfer}, we formulate our problem as follows: given a target task $\mathcal{M}_t \in \boldsymbol{\mathcal{M}}_t$, we aim to learn an optimal policy $\pi^*$ for $\mathcal{M}_t$, by leveraging exterior information from $\boldsymbol{\mathcal{M}}_s$ as well as interior information from $\mathcal{M}_t$.

\section{Recommender system: encoder and policy}
\label{sec:recommender_system}

In this section, we present the architecture of our recommender system.
A frequently used approach to solve POMDPs is to leverage information from past observations in order to build an estimate of the current state, i.e to find a function $\phi \colon \mathbf{o}_1 \dots \mathbf{o}_t \mapsto {\mathbf{z}}_t$ such that $\mathbf{z}_t$ is a good representation of $\mathbf{s}_t$ \citep{moerland2023model}.
Then this estimation is used to take the next action with a function $\psi \colon {\mathbf{z}}_t \mapsto \mathbf{a}_t$.
$\phi$ acts as an encoder and $\psi$ acts as a policy.
We define our recommender system as a composition of the two functions: $\mathcal{F} = \psi \circ \phi$.

In educational contexts, ${\mathbf{z}}_t$ is often taken as an estimate of the student's knowledge -- usually referred to as the \textit{knowledge state} \citep{bassen2020reinforcement, reddy2017accelerating}. 
This knowledge state is traditionally built upon the space of KCs.
However, these KCs are difficult to infer in practice.
Therefore, following the idea from \cite{vassoyan2023towards}, we have built our estimate upon a space of \textit{concepts} (or \say{keywords}).
In this context, a \say{keyword} is a word or group of words that refers to a technical concept closely related to the academic topic of the corpus.
These are much easier to extract than KCs, and therefore better suited for our generalization purposes (we detail in Appendix \ref{appendix:keyword_extraction} how we automatically extract keywords from corpora using an LLM).
Consequently, we model the student's knowledge state as a collection of keyword vectors
${\mathbf{z}}_t = (\mathbf{h}_{w_1}, \mathbf{h}_{w_2}, \dots)$ intended to capture his knowledge of each concept (see Figure \ref{fig:student_keywords} in the appendix) \footnote{Note that even though our experimental pipeline uses KCs to simulate students and properly evaluate the models, in practice our recommender system does not rely on them and makes recommendations solely based on keyword embeddings and human interactions.}.

Regarding the architecture of $\phi$ and $\psi$, note that our framework requires a flexible model that must be able to run on multiple corpora (with different sizes).
Graph neural networks (GNN) have long been identified as an efficient way to make recommendations on flexible data structures: indeed, their number of parameters does not depend on the size of the support \citep{wang2019neural}.
Therefore we define $\phi$ and $\psi$ as GNNs operating on a bipartite graph of documents and keywords $\mathcal{B}=\bigl( (\mathcal{V}_{\mathcal{D}}, \mathcal{V}_{\mathcal{W}}), \mathcal{E}_\mathcal{B}, (E,F) \bigr)$,  where $\mathcal{V}_{\mathcal{D}}$ is the set of documents, $\mathcal{V}_{\mathcal{W}}$ is the set of keywords and $\mathcal{E}_\mathcal{B}$ the set of edges with $(d,w) \in \mathcal{E}_\mathcal{B}$ if the document $d$ contains the keyword $w$.
Regarding the node features, we have used pre-trained word embeddings $E$ for the keyword nodes and the student's past feedback $F$ for the document nodes\footnote{Actually for the document nodes, we have used a combination of keyword embeddings and student's past feedback: more details are provided in Appendix \ref{appendix:recommender_system}.}.
Eventually, for the layers of the GNNs, we have used the graph transformer operator from \citet{ijcai2021p214}:
\begin{multline}
    \mathbf{h}_i^{(l+1)}=\mathbf{W}_1 \mathbf{h}_i^{(l)}+\sum_{j \in \mathcal{N}(i)} \alpha_{i, j} \mathbf{W}_2 \mathbf{h}_j^{(l)} 
    \quad \text{with}\quad \alpha_{i, j}=\operatorname{softmax}\Bigl(\frac{\bigl(\mathbf{W}_3 \mathbf{h}_i^{(l)}\bigr)^{\top}\bigl(\mathbf{W}_4 \mathbf{h}_j^{(l)}\bigr)}{\sqrt{n}}\Bigr)
    \label{eq:transformer_conv}
\end{multline}
where $\mathbf{h}^{(l)}_i \in \mathbb{R}^{n}$ is the embedding of node $i$ at layer $l$ ($h_i^{(0)}$ are the features of the nodes), $\W_1,\W_2,\W_3,\W_4 \in \mathbb{R}^{n\times n}$ are the weight matrices and $\alpha_{i,j} \in \mathbb{R}$ are the attention coefficients, computed via multi-head dot product attention.
This model is similar to the one provided by \citep{vassoyan2023towards}.

\begin{figure}
    \centering
    \includegraphics[width=0.72\columnwidth]{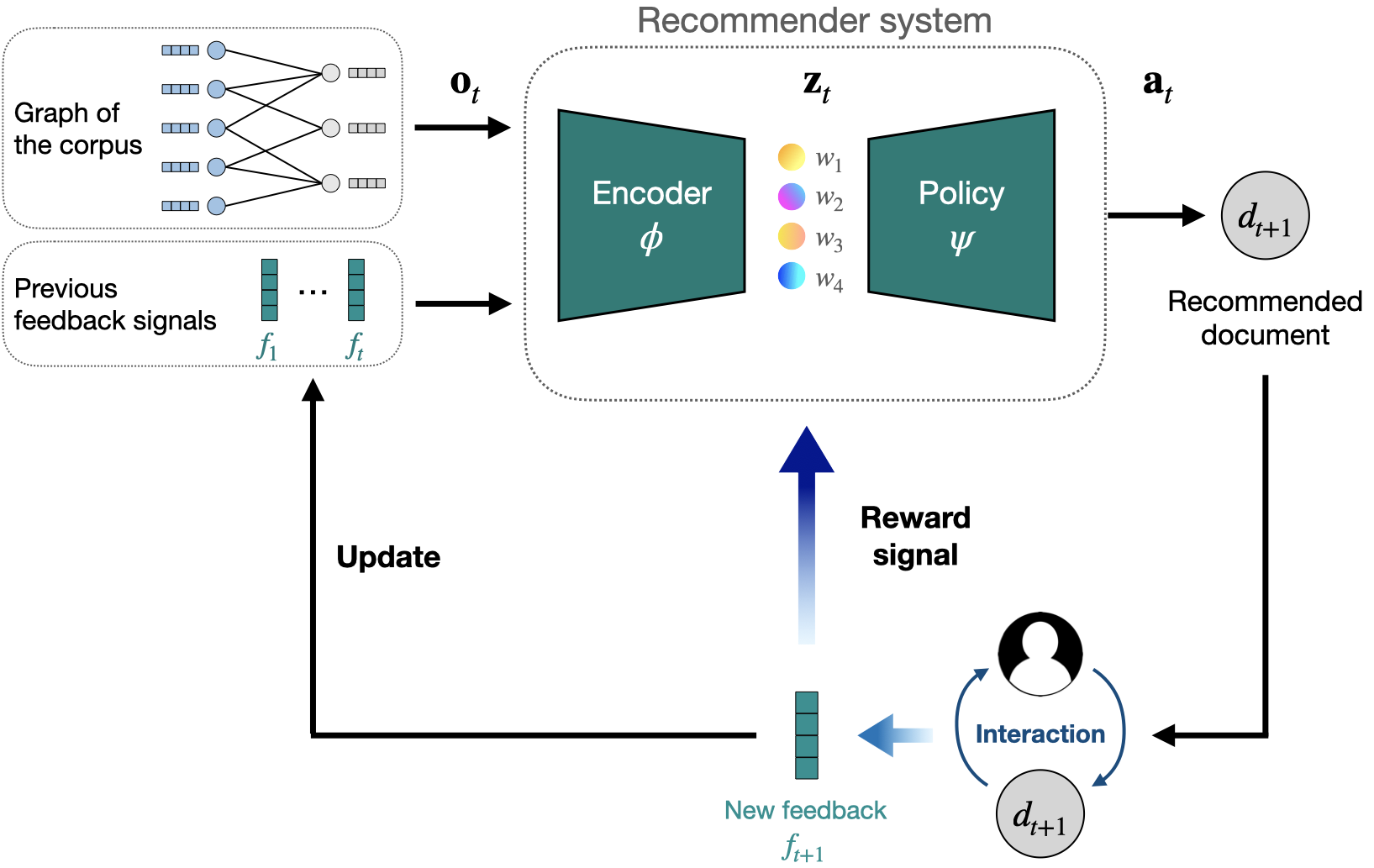}
    \caption{Overview of the recommendation pipeline on one student}
    \label{fig:recommendation_pipeline}
\end{figure}

Note that in the end, there are two graphs: the environment graph \mbox{$G = \bigl(\mathcal{V}, \mathcal{E}, \mathbf{x} \bigr)$} that determines the students behaviour (unknown to the recommender system) and the bipartite graph $\mathcal{B}=\bigl( (\mathcal{V}_{\mathcal{D}}, \mathcal{V}_{\mathcal{W}}), \mathcal{E}_\mathcal{B}, (E,F)\bigr)$ that is passed as an input to the recommender system.
A global view of the recommendation pipeline is provided in Figure \ref{fig:recommendation_pipeline}.
More details about the architecture of the model and its hyperparameters can be found in Appendix \ref{appendix:recommender_system}.

\section{Pre-training on source tasks}
\label{sec:pretraining}

In this section, we present our approach for pre-training our recommender system on sequential corpora.


\subsection{Data distribution}\label{subsec:data_distrib_pretrain}

\paragraph{Corpus}
For the corpus distribution, we have used 14 real-world \textbf{sequential} corpora scraped from 3 popular e-learning platforms.
Each corpus contains a sequence of videos that are meant to be viewed one after another, as they build up from each other.
These corpora address educational topics related to machine learning, statistics, and computer science. 
We automatically extracted keywords from the transcriptions of these videos using an LLM, as described in Appendix \ref{appendix:keyword_extraction}.
For the keyword features, we have used pre-trained embeddings from \textsf{Wikipedia2Vec} \citep{wikipedia2vec}.

\paragraph{Population}
For a sequential corpus, $\mathcal{E}_{\text{pref}} = \varnothing$.
Therefore the only degree of freedom is the distribution over prior knowledge $P(\mathbf{x} | \mathcal{V}, \mathcal{E}_{\text{prereq}})$ with $\mathcal{E}_{\text{prereq}}= k_1 \to k_2 \cdots \to k_N$.
For this pre-training task, we assumed a population with zero prior knowledge for simplicity.
This means that for each student, $\x_{0} = (0 \ldots 0)$ at the beginning of the learning session.

\subsection{Pre-training}

The pre-training was carried out in two stages.
In the first stage, we trained the model to match the predictions of an oracle, in a supervised learning way.
Our oracle is simply an algorithm that has access to the environment graph and therefore knows exactly which document to recommend in each situation.
Note that this first stage is quite similar to the way LLMs are pre-trained: instead of predicting the next token, we predict the next document in a predefined sequence.
In the second stage, a reinforcement learning training was performed with the \texttt{REINFORCE} algorithm \citep{sutton1999policy}.
This stage is crucial as, with a discount factor greater than zero, the model can learn to make \say{useful} mistakes
—that is, recommendations that do not facilitate student learning but provide valuable information about his current knowledge.
The choice of not using any algorithm involving an approximate value function, although these are known to be more stable \citep{mnih2013playing, haarnoja2018soft, mnih2016asynchronous}, was motivated by the difficulty for GNNs to reason at the scale of the whole graph (which is required for a good estimate of the state value).
We directly trained our RL agent on the collection of 14 POMDPs induced by our sequential corpora.
The hyperparameters we have used for this pre-training stage as well as the corresponding learning curve are provided in Figure \ref{fig:source-task-results} of the Appendix.
We have trained the RL agent on approximately 25k steps.
The maximum length of each episode was set according to the number of documents in the current corpus.
Other pre-training strategies have also been tested, but with weaker results.
Results and discussion on these are provided in Appendix \ref{appendix:other-pretraining}.

\section{Fine-tuning on target task}
\label{sec:fine_tuning}

\subsection{Data distribution}\label{subsec:fine-tuning-data-distrib}

\paragraph{Corpus}
To evaluate our recommender system in a complex adaptive learning scenario, we have designed a corpus of 22 written documents teaching machine learning basics.
We have conceived it in a way that the dependencies between knowledge components are easy to establish manually.
The resulting graph is a grid of KCs with 3 rows and 11 columns, depecting prerequisite relationships among all documents.
We have also included an additional \say{background} KC that isn't taught by any document but conditions the access to some of them, thus bringing greater diversity to the student population.
More details about the design of this corpus along with a graphical representation are provided in Appendix \ref{appendix:graph_corpus}. 

\paragraph{Population}
In our framework, choosing a student distribution is equivalent to defining $P(\Epref | \C)$ and $P(\x | \Epref, \C)$.
Regarding $P(\Epref | \C)$, we considered that in the specific case of our corpus, it made sense to model learning preferences as a set of extra \say{vertical} dependencies (edges) between knowledge components (see Figure \ref{fig:graph_corpus} of the Appendix).
Therefore, for each new student, a random set of such edges was generated, following a binomial distribution of parameter $p=0.3$.
So for any $k_{i,j}\in \mathcal{V},  P(k_{i,j}\to k_{i+1,j} \in \Epref)=0.3$.
An interpretation of this modeling of learning preferences is provided in Appendix \ref{appendix:graph_corpus}.
As for the distribution over prior knowledge $P(\x | \Epref, \C)$, we have considered 3 possible scenarios:
\begin{description}
    \item Scenario 1 (none): the learners have no prior knowledge, i.e. $\x = (0 \ldots 0)$ for all students;
    \item Scenario 2 (decreasing exponential): the number of KCs known by the students prior to the learning session follows a decreasing exponential distribution; this allows to model a population where most students have little knowledge about the corpus, but not all of them;
    \item Scenario 3 (uniform): the number of KCs known by the students prior to the learning session follows a uniform distribution; this is the most challenging environment as uncertainty is maximal.
\end{description}

\subsection{Training}

For each sampled student, some paths in the corpus are reachable while some others are not (depending on their prior knowledge and learning preferences).
To avoid that the agent always chooses the safest path (instead of adapting to the student's profile), we slightly modified the reward function, in order to give greater rewards to \say{difficult} paths.
Rather than summing the newly acquired knowledge components as depicted in Equation \ref{eq:reward}, we computed a weighted sum by multiplying each KC by its respective value.
More details about the values of the KCs are provided in Appendix \ref{appendix:graph_corpus}.

The fine-tuning was performed on 10 epochs, where 5 new students were sampled on each epoch.
Therefore, it was carried out on a total of 50 students.
The maximum size of each episode was set to 11 (because of the 11 \say{major} concepts, cf. Appendix \ref{appendix:graph_corpus}). 
On each epoch, we tested our model on 20 test episodes ($\sim$ 20 students) and aggregated our results over 30 random seeds.
We did not perform any hyperparameter search in the fine-tuning stage so as not to give an advantage to our pre-trained model.
Instead, we simply adopted the hyperparameters used by \citet{vassoyan2023towards} as they provided good performance for training on a single corpus (reported in Table \ref{table:hyperparams_rl_finetune} of the Appendix).

\begin{figure*}[htbp]
    \centering
    \includegraphics[width=0.99\textwidth]{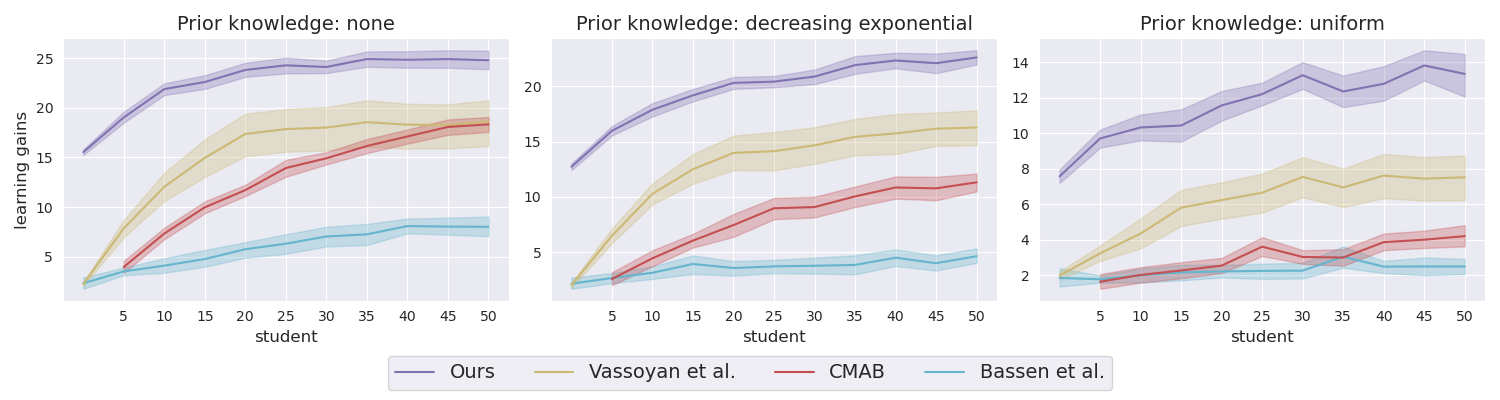}
    \caption{Performance comparison (mean and confidence interval) depending on the pre-training strategy, for multiple prior knowledge distributions in the population. At each epoch, new data are collected from 5 new sessions with simulated students. In total, the training is carried out on a group of 50 students. These results are aggregated over 30 random seeds.}
    \label{fig:results_fine_tuning}
\end{figure*}

\begin{table*}[htbp]
\centering
\medskip
\begin{tabular}{c|cccc}
\toprule
 {Prior knowledge} & {Ours} & {Vassoyan et al.} & {CMAB} & {Bassen et al.}  \\
\midrule
None              & \textbf{24.81\ (2.63)} & 18.63\ (6.55) & 18.34\ (2.11) & 8.02\ (1.62)  \\
Decreasing exp.    & \textbf{22.62\ (1.82)} & 16.28\ (4.51) & 11.32\ (2.24) & 4.64\ (1.07)  \\
Uniform           & \textbf{13.33\ (3.40)} & 7.51\ (3.44)  & 4.19\ (1.68)  & 2.49\ (0.70) \\
\bottomrule
\end{tabular}
\caption{Final episodic returns (average and standard deviation) on each task.}
\label{tab:target-task-results}
\end{table*}

\subsection{Baselines}

Although numerous approaches had been explored in the literature on learning path personalization, for a fair comparison we selected only models that did not use the corpus's prerequisite graph for making predictions.
Consequently, we compared our approach against three baselines.

First is the approach from \citet{bassen2020reinforcement}, which was used to recommend a sequence of educational activities to students.
Their model consists of 2 fully-connected neural networks (one for the actor and one for the critic) trained with Proximal Policy Optimization \citep{schulman2017proximal}.
Since our learning scenario is quite different from theirs, we had to adapt the observation space: in particular, the pre-test scores and the scores on activities were omitted.
Instead, we just kept a feature vector stating which feedback was given on each document by the current student.
The authors did not provide the hyperparameters they used in their experiment.
Therefore, we carried out a hyperparameter search over various learning rates, layer sizes, batch sizes and \say{repeat per collect} \footnote{This parameter specifies the number of times the policy is updated for each batch of data collected. For instance, setting it to 2 indicates that the policy will undergo two learning iterations for each batch of collected data.} values and kept the best.

Our second baseline was the linear contextual multi-armed bandit (CMAB), with Thompson Sampling as the policy \cite{agrawal2013thompson}.
The bandit requires context for each action instead of states, which we provided as the number of times the document was previously recommended to the student, and the number of times it was \say{understood}.

Finally, we compared our approach against the model from \citet{vassoyan2023towards}, which is very similar to ours, except that it does not feature any type of pre-training.

\subsection{Results and discussion}

The results of our fine-tuning experiments are presented in Figure \ref{fig:results_fine_tuning} and Table \ref{tab:target-task-results}.
The four recommendation engines are compared in each of the three scenarios described in \ref{subsec:fine-tuning-data-distrib}.
The performance of each model is measured in learning gains per student, which, in our setting, is equivalent to the undiscounted episodic return.
These learning gains are plotted against the number of students, where each student can be regarded as an episode.
The maximum episode length is 11, which puts us in a very low data regime. 
The margins of error in Figure \ref{fig:results_fine_tuning} were estimated using a studentized bootstrap method, with 10,000 resamples and a 95\% confidence interval.

The first observation that can be made from these plots is that our pre-trained model outperformed all other approaches in all scenarios.
As expected, the performance gap was most pronounced at the beginning of the training sessions, where our model benefited from a strong warm start.
Moreover, it maintained its lead throughout the entire training sessions and converged to values well above those of its non-pre-trained counterparts.
This suggests that the pre-training process has given our model some notion of dependencies among keyword embeddings, helping it to quickly find good recommendation strategies, even though it had never previously encountered the target corpus.

Surprisingly enough, our model seems to perform just as well in scenarios 2 and 3, despite not having been exposed to them in the pre-training stage.
This shows that in the pre-training stage, the model has acquired insights that extend beyond mere knowledge of the population's distribution.
In comparison, a simple baseline like CMAB performed quite well when the population had no prior knowledge but struggled more and more as the uncertainty increased.
The model from \citep{bassen2020reinforcement} got the worst performance, despite being the only one that benefited from a hyperparameter search on this corpus.
One explanation could be that this model was designed for a real world setting with fewer educational activities and much more students ($\sim 1000$).

Eventually one can note from Table \ref{tab:target-task-results} that despite being trained with the \texttt{REINFORCE} algorithm, which, in comparison to PPO, usually suffers from high variance, our model showed a relatively low variance (much lower than its non-pre-trained counterpart).

\section{Related work}
\label{sec:rw}

Previous works have used reinforcement learning to decide on the pedagogical action in computer-based education to increase the student's learning gain.
Many works used RL agents with small state and action spaces to accommodate the small amount of data available.
For such cases, a tabular RL algorithm was used to train an RL agent.
One example of this is the work by \citet{chi2011empirically}, where the RL agent takes a state vector containing e.g. the simplicity of the concept to explain, and decides between hinting and telling the student how to solve an exercise.
Similar approaches have been shown to work in practice in specific and tightly constrained settings \citep{gordon2016affective,park2019model}.
Because of the simplicity of the state space, these works are limited to small and fixed, specific sets of interactions.

Another branch of approaches used simulated students to allow for more learning iterations.
Among the first was the work by \citet{iglesias2009}, where tabular Q-learning was used to choose the types of learning material for simulated students.
Later works have used RL algorithms that can handle more complex state and action spaces \citep{lan2016contextual,rafferty2016faster}, and in particular deep RL \citep{reddy2017accelerating,subramanian2021deep}. 
Although simulated students allow RL agents to train for virtually unlimited episodes, and so handle more complex dynamics, there is a strong reliance on the accuracy and realism of the simulated students. 
We alleviate this by using simulated students for pre-training, then evaluate the potential of the RL agent to generalize to an unseen learning environment.

Closely related to our work was by \citet{bassen2020reinforcement}, which applied deep RL to recommend different educational materials from a pool of 12.
They used data collected from 1{\small,}830 participants to train the agent. 
They used one-hot encoding to represent the materials, meaning that the agent needs to be retrained to work with another set of materials.
Their approach also required a large number of real student data because of the sparse reward, coming only once at the end of an episode, showing the importance of sample efficiency in real-world applications.
We improve upon this by using meaningful embeddings to represent the materials to recommend.

\ifx\EAAI\undefined
\else

Designing simulated students requires modeling how knowledge works. A student's knowledge is broken down into Knowledge Components (KCs), which are small units of knowledge that a student can learn and hold (e.g. a fact or a skill).
Based on this, different exercises or content can be associated with different KCs, creating a binary matrix called the q-matrix \citep{barnes2005q}, allowing interactions with an exercise to act as evidence for the associated KCs.
Building up further, the theory of knowledge space and the attribute hierarchy method both suggest an order of prerequisites between these KCs, which can be represented as a graph \citep{doignon1985spaces,leighton2004attribute}.

\fi

\section{Conclusion}

Our study shows that when dealing with an adaptive learning scenario, one efficient strategy to improve sample-efficiency is to gather a set of closely related sequence corpora and pre-train a recommender system on them. This pre-training procedure does not involve any annotator to tag documents nor real students to interact with them, but requires a flexible enough recommender system. This opens new avenues for applications of deep RL algorithms to personalized learning environments, and paves the way for the developement of scalable and reusable models for adaptive learning. In the future, this method could be tested on a wider range of adaptive learning scenarios as well as on real-world student populations.

\bibliographystyle{plainnat}
\bibliography{aaai25}

\appendix
\onecolumn

\section{Details about the environment}\label{appendix:details_env}




\subsection{Learning preferences}\label{appendix:learning_prefs}

We argue that learning preferences can be modelled without loss of generality through a set of additional dependencies between knowledge components (if necessary by adding more knowledge components).
To illustrate this point, let's consider two students $u_1$ and $u_2$ mastering the same set of knowledge components: $\mathbf{x}_{u_1} = \mathbf{x}_{u_2} = \mathbf{x}$.
Let's suppose that $u_1$ and $u_2$ hold different learning preferences: this means that there exists at least one document $d$ for which $u_1$ and $u_2$ will react differently, for example $u_1$ understands it while $u_2$ does not.
Assuming that $d$ is not definitively out of scope for student $u_2$, there must exist a minimal set of documents $\mathcal{D}^\prime$ whose reading will enable him to understand document $d$.
Denoting $\mathcal{K} = d_\rightarrow$ and $\mathcal{K}^\prime = \cup_{d^\prime \in \mathcal{D}^\prime} d^\prime_\rightarrow$, this translates into an extra set of dependencies $\mathcal{K}^\prime \rightarrow \mathcal{K}$ for student $u_2$.

\subsection{Transition probability function}\label{appendix:transition-probability}


To describe the transition probability of the student's dynamics, we start with the base case, where there is only one KC in the dynamics. We model the student's knowledge state as a binary ``known'' or ``unknown'' state on the KC.
The student never forgets, so a known KC remains known for the rest of the episode.
Learning can occur when a student interacts with a document that teaches a KC that the student previously did not know.
If the student knows all the prerequisite KCs required to understand the document, then the student will learn all the KCs taught by that document.
Otherwise, the student will not learn any of the KCs taught.
As a result, the transition of the student's state is fully deterministic, dependent on the stochastic and unknown prerequisites of the documents.
Formally, the transition probability in the case of 1 KC is described as the following: $\tau(\x, d, x_i^\prime)  = \mathbb{P} (\mathbf{s}_{t+1}^{(i)} = x^{\prime}_i \mid \mathbf{s}_{t}=\x, \ \mathbf{a}_{t}=d)$:
\begin{equation}
    \tau(\x, d, x_i^\prime) = 
\begin{cases}
    1, & \text{if} \quad (x_i=1, \ x_i^\prime=1)\\
    0, & \text{if} \quad (x_i=1, \ x_i^\prime=0)\\
    \mathds{1}(k_i\in d_\rightarrow )\ \sigma\left(\sum\limits_{k_j \in d_\leftarrow} \mathds{1} (x_j=0) \right)\\ + \mathds{1}(k_i \notin d_\rightarrow), & \text{if} \quad (x_i=0, \ x_i^\prime=0)\\
    \mathds{1}(k_i\in d_\rightarrow )\prod\limits_{k_j \in d_\leftarrow} \mathds{1} (x_j=1), & \text{if} \quad (x_i=0, \ x_i^\prime=1)\\
\end{cases}
\label{eq:transition-verbose}
\end{equation}
Through straightforward manipulations, we can transform the four cases of Equation \ref{eq:transition-verbose} into Equation \ref{eq:transition-short}.

We then extend the student's dynamics to work with multiple KCs, which we build up from the base case.
For this, we model the learning of the different KCs independently, meaning that the transition of one KC does not affect how another KC is learned.
Using the notations from Section \ref{sec:formulation}, the full transition probability function on multiple KCs is:
\begin{equation}
\mathbb{P}(\mathbf{s}_{t+1} = s^\prime \mid \mathbf{s}_{t}=s, \ \mathbf{a}_{t}=d ) = \prod_{k_i \in \mathcal{V}} \mathbb{P} (\mathbf{s}_{t+1}^{(i)} = s^{\prime}_i \mid \mathbf{s}_{t}=s, \ \mathbf{a}_{t}=d)
\end{equation}

\subsection{Observation function}

Next is the observation function, which describes how a student responds to a recommended document.
There are three possible interactions for the student: right level, too easy, or too hard, denoted $f_\circ$, $f_>$, $f_<$ respectively.
The student deems the document too easy when the student has already mastered the KCs that the document teaches.
In contrast, the student deems the document too hard when the student did not master all the prerequisite KCs of the document at the time.
Otherwise, the document is at the right level, making the student master new KC(s).
Formally, the observation function $\mathcal{Z}(s, d, o)$ is described as follows, where $s = (\x, \Epref)$:

\begin{equation}
    \mathcal{Z}(s, d, o) = 
\begin{cases}
    \mathds{1}(\x \nsupseteq d_\leftarrow), & \text{if} \quad (o = f_<)\\
    \mathds{1}(d_\rightarrow \subseteq \x), & \text{if} \quad (o = f_>)\\
    \mathds{1}(\x \supseteq d_\leftarrow \wedge x_i \cap d_\rightarrow \neq x_i), & \text{if} \quad (o = f_\circ)\\
\end{cases}
\end{equation}

\begin{figure}
    \centering
    \includegraphics[width=0.55\columnwidth]{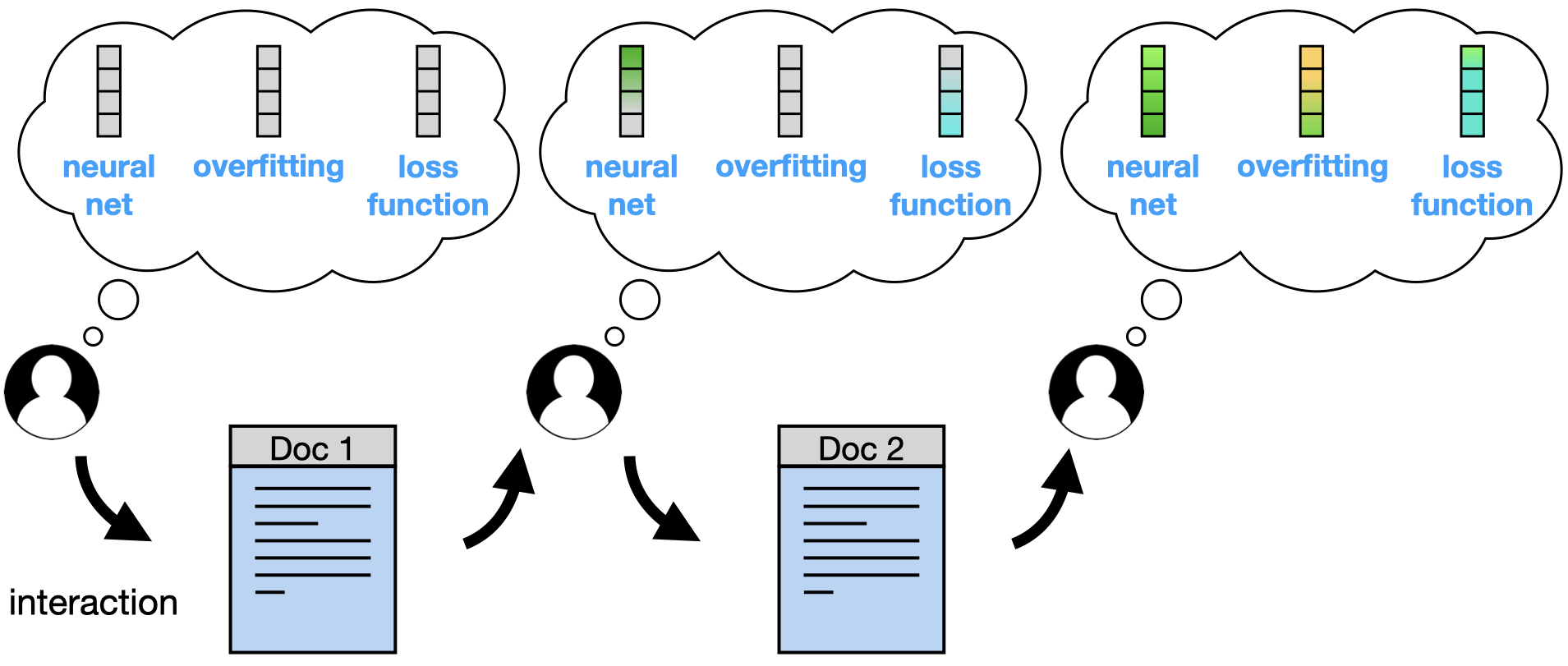}
    \caption{An intuitive view of student's knowledge state model with keyword vectors}
    \label{fig:student_keywords}
\end{figure}

\section{Architecture of the recommender system}\label{appendix:recommender_system}

Our recommender system $\mathcal{F} = \psi \circ \phi$ is a composition of two graph neural networks operating on the bipartite graph $\mathcal{B}$ defined in Section \ref{sec:recommender_system}.
GNN layers are implemented with directed edges, successively used from documents to keywords, and from keywords to documents.
Moreover, $\phi$ combines a GNN architecture with a multi-layer perceptron (MLP) that produces embeddings of the feedback signals.
These feedback embeddings are then combined with the embeddings of their associated document nodes with a Hadamard product.

Below is the complete architecture of our recommender system:
\begin{align}
    \begin{rcases}
    H^{(1)} &= \textsf{Linear}(E)  \\
    H_W^{(2)} &= \textsf{TransformerConv}_{\text{doc} \rightarrow \text{kw} }(H^{(1)}) \qquad \\
    H_D^{(2)} &= \textsf{TransformerConv}_{\text{kw} \rightarrow \text{doc} }(H^{(2)})  \\
    H_D^{(3)} &= H_D^{(2)} \odot \textsf{MLP}(F_D)\\
    H_W^{(3)} &= \textsf{TransformerConv}_{\text{doc} \rightarrow \text{kw} }(H^{(3)})
    \end{rcases} \phi \\
    \begin{rcases}
    H_D^{(4)} &= \textsf{TransformerConv}_{\text{kw} \rightarrow \text{doc} }(H^{(3)}) \qquad \\
    Z_D &= \textsf{Linear}(H_D^{(4)})
    \end{rcases}\psi
\end{align}

where $H^{(l)} \in \mathbb{R}^{|\mathcal{V}|\times k}$ is the node embeddings at layer $l$, with $H_D^{(l)} \in \mathbb{R}^{|\mathcal{V}_D|\times k}$ and $H_W^{(l)} \in \mathbb{R}^{|\mathcal{V}_W|\times k}$ are the restrictions of $H^{(l)}$ to document and keyword nodes respectively.
Note that $H_W^{(3)}$ is actually the latent representation $\mathbf{Z}_t$ defined in section \ref{sec:recommender_system}.
$E \in \mathbb{R}^{n\times n}$ is the input node features.
$F_D$ is the matrix of past feedback signals (one feedback per document, with a \say{none} feedback for documents that have not been visited yet).
$\textsf{TransformerConv}(\cdot)$ is the operator formulated in Equation \ref{eq:transformer_conv}, with the subscripts $\text{doc} \rightarrow \text{kw} $ and $\text{kw} \rightarrow \text{doc}$ denoting the directions of edges.
$\textsf{Linear}(\cdot)$ is a linear transformation.
$\textsf{MLP}(\cdot)$ is a two layers perceptron with one hidden layer.
$\odot$ is the Hadamard product.
An extra linear layer is added after the final \textsf{TransformerConv} to map document representations to scores (then converted to probability distribution via softmax).

An exhaustive list of the hyperparameters used in our model is provided in Table \ref{table:hyperparams_model}.

\begin{table}[H]
    \centering
    \medskip
    \begin{tabular}{ccc}
        \toprule
        Hyperparameter & Value\\ \midrule
        Hidden dimension                       & $128$  \\
        Activation function                    & Exp. Linear Unit (ELU)   \\
        Attention type                         & Additive \\
        Number of attention heads              &  $4$   \\
        \textsf{Wikipedia2Vec} embedding size & $100$ \\
        \bottomrule
    \end{tabular}
    \caption{Hyperparameters of the recommender system}
    \label{table:hyperparams_model}
\end{table}

\section{Pre-training on source tasks}

\begin{figure}[t]
  \centering
  \begin{subfigure}[c]{0.4\columnwidth}
    \centering
    \includegraphics[width=\linewidth]{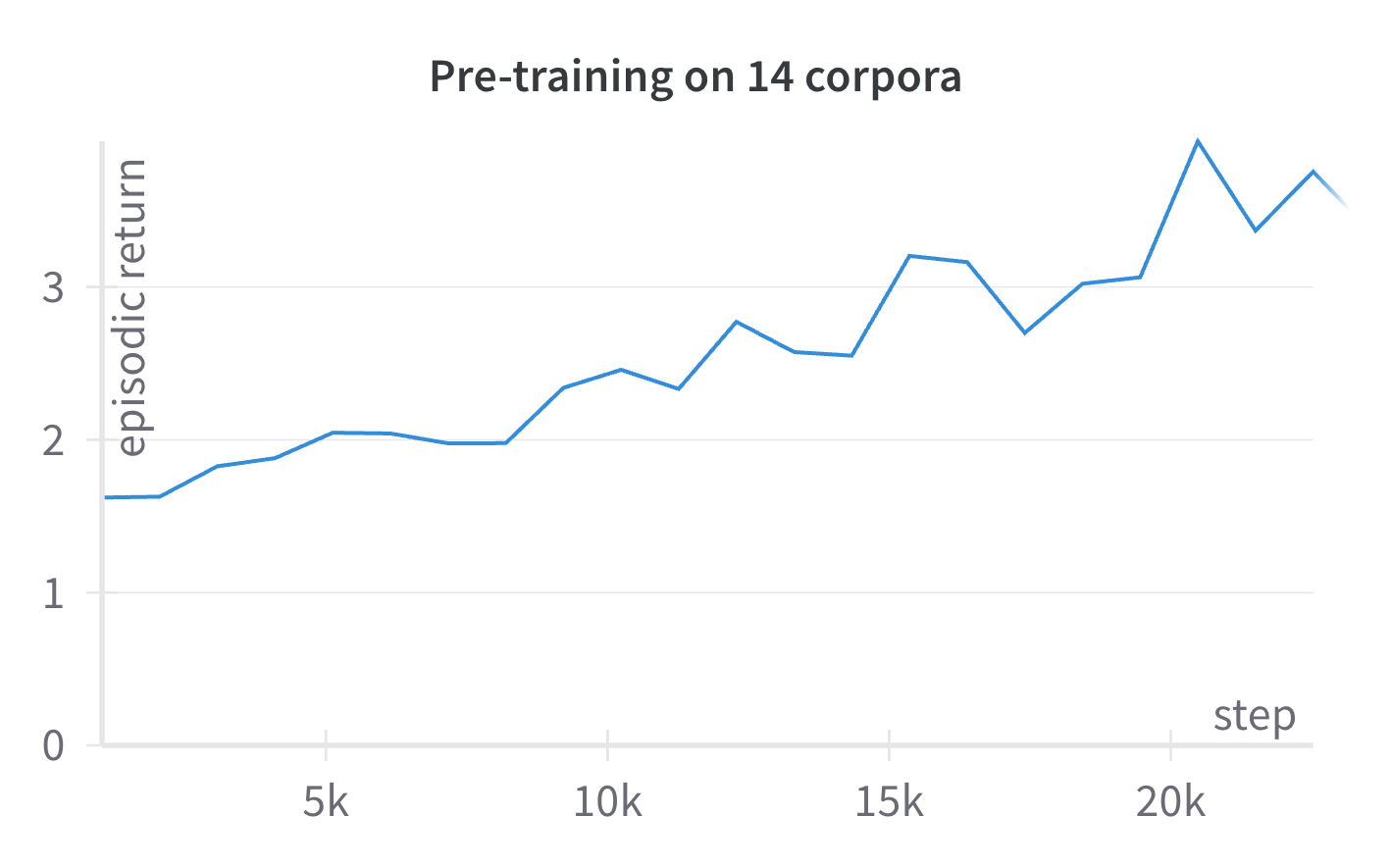}
  \end{subfigure}
  \begin{subfigure}[c]{0.45\columnwidth}
    \centering
        \begin{tabular}{cc}
            \toprule
            Hyperparameter & Value\\ \midrule
            Discount factor                        & 0.7 \\
            Learning rate                          & 0.0001  \\
            Entropy coefficient                    & 0.01 \\
            Batch size                             &  8  \\
            Repeat per collect                     &  15  \\
            Steps per collect                      &  1024  \\
            \bottomrule
        \end{tabular}
  \end{subfigure}
  \caption{Learning curve and hyperparameters used for the RL pre-training stage}
  \label{fig:source-task-results}
\end{figure}

The learning curve and hyperparameters we have used for the pre-training on source tasks are provided in Figure \ref{fig:source-task-results}.
The batch size refers to the number of graphs passed as input to the model (hence the small value).

\section{Fine-tuning on target task}\label{appendix:fine-tuning-target-task}

\subsection{Graph corpus}\label{appendix:graph_corpus}

The graph corpus contains a set of written documents on machine learning, which are distinct from the ones in the sequential corpora, albeit they share some keywords.
We have designed this corpus in such a way that it is aimed at two types of student: those with Computer Science (CS) background, and those without CS background.
Consequently, the corpus teaches 11 major concepts and each concept is addressed by 2 documents, from 2 different angles: the \say{computer science} angle (often involving linear algebra or IT concepts) and the \say{non-computer science} angle.
To model this, we have adopted a 3-line system of knowledge components, where each major concept is broken down into three knowledge components (one at each level): the middle KC represents the major concept to be taught, the top one represents its "non-CS" angle and the bottom one represents its "CS" angle.
Consequently, all the bottom KCs are conditioned by a \say{background} KC, which states whether the student has a \say{CS} background or not.
Non-CS students can only learn from non-CS documents, as they lack the CS prerequisite.
CS students can learn from both types of documents.
Prerequisites relationships $\Eprereq$ were modeled as horizontal dependencies in the two bottom levels.
As for learning preferences $\Epref$, these were modeled as \say{vertical} dependencies.
Here is an example to motivate this choice: consider that one major concept of the corpus is the artificial neuron.
It can be taught using CS notions (like linear algebra) or without these notions (for instance using a biological analogy).
For some CS students, the mathematical explanation may be sufficient to enable them to understand the concept intuitively.
But for some others, although they have the necessary CS background, some non-mathematical explanation might help in grasping the concept intuitively.
Hence the existence of certain vertical dependencies between the \say{non-CS} and \say{CS} levels, which we model as learning preferences.
Overall, the graph corpus aims to show that the RL agent can generalize its knowledge from the sequential corpora, and that the RL agent can learn to adapt to different types of students.

The \say{value} of each KC depends on its position on the grid: the value is 1 for row 1, 2 for row 2 and 3 for row 3 (the most difficult to access are the most valuable).

\begin{figure}[H]
    \centering
    \includegraphics[width=0.8\columnwidth]{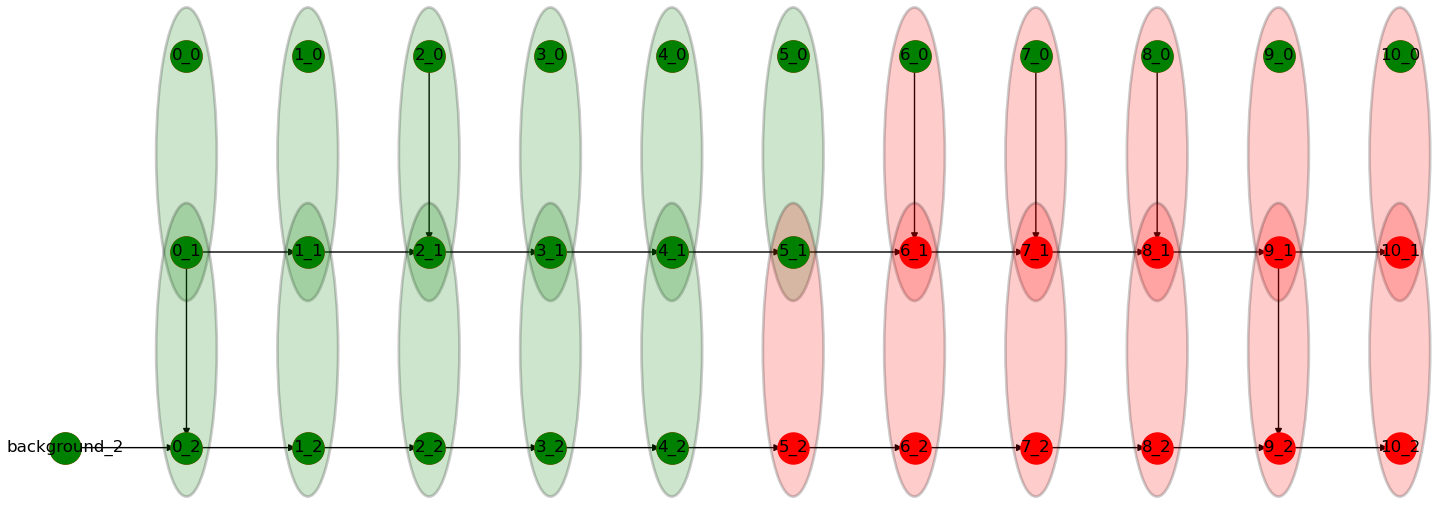}
    \caption{Illustration of the graph corpus we have used for the fine-tuning task. The nodes are the knowledge components and the ellipses that enclose them are the documents. This graph actually illustrates the sampling of a student, as the green color represents the prior knowledge of the student and the vertical edges represent his learning preferences. }
    \label{fig:graph_corpus}
\end{figure}

\subsection{Fine-tuning process}

The hyperparameters we have used for the fine-tuning of the model on the target task are listed in Table \ref{table:hyperparams_rl_finetune}.

\begin{table}[H]
    \centering
    \medskip
    \begin{tabular}{ccc}
        \toprule
        Hyperparameter & Value\\ \midrule
        Learning rate                          & $0.0005$  \\
        Batch size                             &  16  \\
        Repeat per collect                     &  15  \\
        Episodes per collect                   &  5  \\
        Discount factor                        & $0$ \\
        \bottomrule
    \end{tabular}
    \caption{Hyperparameters used for the RL fine-tuning}
    \label{table:hyperparams_rl_finetune}
\end{table}

\section{Other pre-training strategies}\label{appendix:other-pretraining}

\begin{figure}[htbp]
    \centering
    \includegraphics[width=0.99\textwidth]{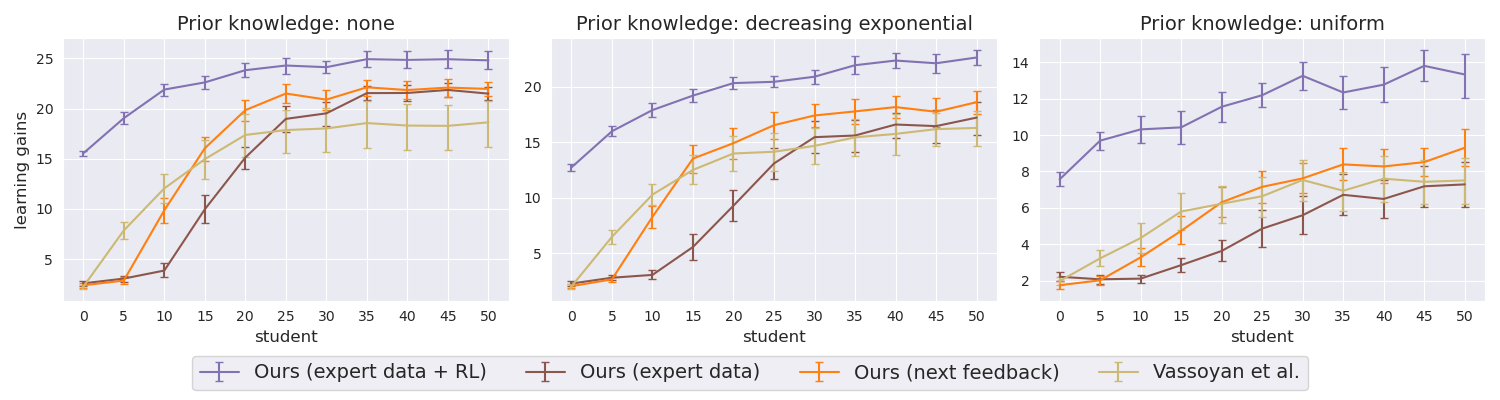}
    \caption{Performance comparison of our model against two other pre-training strategies.}
    \label{fig:results_other_pretraining}
\end{figure}

In this section we provide experimental results for two other pre-training strategies.
Both of them are performed on the same source tasks (sequential corpora) we have used in the main experiment.
The first strategy is just the same as the one we presented in the core of the paper, without the RL stage.
This means the model was solely pre-trained on expert data, allowing to evaluate the importance of the RL stage.
The second strategy consists in training the model to predict student's feedback on each next document (instead of choosing next document).
This provides a target signal slightly more informative than the sole good/bad recommendation signal we have used for pre-training on expert data.
Therefore, this could lead to more accurate internal representations.
The learning curves are presented in Figure \ref{fig:results_other_pretraining} (our final model is also displayed for comparison).
The margins of error were estimated using a studentized bootstrap method, with 10,000 resamples and a 95\% confidence interval.

The first takeaway of this experiment is that these two methods, even if they manage to slightly outperform their non-pre-trained counterpart (Vassoyan et al.), perform much worse than our main model.
This is especially pronounced at the beginning of the fine-tuning.
In the case of the model pre-trained on predicting next feedback, such cold-start is not surprising, as the model is just discovering the recommendation tasks.
However, the model pre-trained on expert data was trained precisely on a recommendation task, and its poor performance at the start of fine-tuning is all the more surprising.
Although this phenomenon is difficult to interpret at this stage, it does show the importance of RL in pre-training, which alone enables to make good recommendations from the start of fine-tuning.




\section{Keywords extraction}\label{appendix:keyword_extraction}

The main difficulty in our keyword-extraction setting is that we did not extract \say{regular} keywords but only keywords related to Wikipedia articles.
Indeed, we have chosen to use \textsf{Wikipedia2Vec} \citep{wikipedia2vec} pre-trained embeddings as features of the keywords, mainly because we consider that embeddings which primarily contain \say{encyclopedic} information would be more suitable for this task.
Since modern large language models usually have a good knowledge of Wikipedia, we simply prompted GPT-4 \citep{achiam2023gpt} to achieve this task, using the prompt below (adapted for each document):

\begin{tcolorbox}[colback=blue!5!white,colframe=blue!75!black,title=System Prompt:]
    You are a helpful assistant who extract keywords from educational documents.
A user will pass an educational document.
First, you extract the academic topic of the document. This could be the title of the document in a course sequence. Don't write anything else.
Then, you examine every single term in this document and extract them if they refer to a technical concept closely related to the academic topic of the document. Avoid including keywords that are only used in examples and don't have much to do with the subject taught in the document. You directly output a python list of strings (for example: \say{machine learning}, \say{neural network} etc.), without duplicate. There should never be more than thirty keywords or so, so choose the most important ones. Again, don’t write anything else.
Eventually, you associate each term with its corresponding Wikipedia page (Wikipedia ID) if it exists. Sometimes this needs to be done cleverly: for example, \say{sequential data} should be associated with the Wikipedia ID \say{Sequence}. Drop the terms that don’t have a Wikipedia page. Directly output a python list of strings of Wikipedia IDs (for example: [\say{Machine\_learning}, \say{Artificial\_neural\_network}, ...]). Again, don’t write anything else.
\end{tcolorbox}

\begin{tcolorbox}[colback=blue!5!white,colframe=blue!75!black,title=Human Prompt:]
    Machine Learning allows computers to make decisions without being explicitly programmed to do so.
Imagine you are teaching a child to recognize different fruits. You might show the child various pictures of apples, bananas, oranges, and so on. Over time, the child learns to identify these fruits correctly. Machine learning works in a similar way. You feed the computer a lot of data—like pictures of fruits—and the computer learns to recognize patterns within that data to differentiate between them. After sufficient learning, it can correctly identify or predict new, unseen data. For example, if you show it a new picture of an apple, it will be able to identify it as an apple.
\end{tcolorbox}

\begin{tcolorbox}[colback=red!5!white,colframe=red!75!black,title=GPT4:]
    The purpose of Machine Learning.
    
["machine learning", "computer", "make decisions", "program", "learn", "data", "patterns", "predict"]

["Machine\_learning", "Computer", "Decision-making", "Computer\_program", "Learning", "Data", "Pattern\_recognition", "Prediction"]
\end{tcolorbox}

\section{Software}

Our GNN was implemented using the Python library Pytorch-Geometric \citep{pyg} and the reinforcement learning agent was trained with Tianshou \citep{tianshou}.
The keyword extraction process was implemented using Langchain \citep{Chase_LangChain_2022}.

\end{document}